\documentclass{l4dc2026}

\title[M2M-DC]{MI-to-Mid Distilled Compression (M2M-DC): An Hybrid--Information-Guided--Block Pruning with Progressive Inner Slicing Approach to Model Compression}
\usepackage{times}
\usepackage{float}
\usepackage{booktabs} 

\usepackage{algorithm}
\usepackage{algorithmic}
\usepackage{array}
\usepackage[most]{tcolorbox} 
\usepackage{xcolor}
\usepackage{lipsum}  
\tcbset{
  callout/.style={
    enhanced,
    breakable,
    colback=gray!10,
    colframe=gray!60!black,
    boxrule=0.9pt,
    arc=8pt,
    left=6mm,right=6mm,top=3mm,bottom=3mm,
    drop shadow=black!25
  }
}
\newtcolorbox{calloutbox}[1][]{callout,#1}

\usepackage[utf8]{inputenc} 

\DeclareUnicodeCharacter{2212}{\ensuremath{-}}

\author{%
 \Name{Lionel M. Levine} \Email{Lionel@cs.ucla.edu}\\
 \addr Department of Computer Science, UCLA
 \AND
 \Name{Haniyeh Ehsani Oskouie} 
 \Email{Haniyeh@cs.ucla.edu}\\
 \addr Department of Computer Science, UCLA
 \AND
  \Name{Sajjad Ghiasvand} 
 \Email{sajjad@ucsb.edu}\\
 \addr Department of Electrical and Computer Engineering, UCSB
 \AND
\Name{Dr. Majid Sarrafzadeh} \Email{Majid@cs.ucla.edu}\\
 \addr Department of Computer Science, UCLA
}

\begin{document}

\maketitle

\begin{abstract}
We introduce \textbf{MI-to-Mid Distilled Compression (M2M-DC)}, a two-scale, shape-safe compression framework that interleaves information-guided block pruning with progressive inner slicing and staged knowledge distillation (KD). First, M2M-DC ranks residual (or inverted-residual) blocks by a label-aware mutual information (MI) signal and removes the least informative units (structured prune-after-training). It then alternates short KD phases with \emph{stage-coherent, residual-safe} channel slicing: (i) stage ``planes'' (co-slicing \texttt{conv2} out-channels with the downsample path and next-stage inputs), and (ii) an optional mid-channel trim (\texttt{conv1} out / \texttt{bn1} / \texttt{conv2} in). This targets complementary redundancy—whole computational motifs and within-stage width—while preserving residual shape invariants.

On CIFAR-100, M2M-DC yields a clean accuracy--compute frontier. For \textbf{ResNet-18}, we obtain \textbf{85.46\%} Top-1 with \textbf{3.09M} parameters and \textbf{0.0139} GMacs (\(-72\%\) params, \(-63\%\) GMacs vs.\ teacher; mean final \(85.29\%\) over three seeds). For \textbf{ResNet-34}, we reach \textbf{85.02\%} Top-1 with \textbf{5.46M} params and \textbf{0.0195} GMacs (\(-74\%/-74\%\) vs.\ teacher; mean final \(84.62\%\)). Extending to inverted-residuals, \textbf{MobileNetV2} achieves a mean final \textbf{68.54\%} Top-1 at \(\sim 1.71\)M params (\(-27\%\)) and \(\sim 0.0186\) conv-GMacs (\(-24\%\)), improving over the teacher's \(66.03\%\) by \(+2.5\) points across three seeds.

Because M2M-DC exposes only a thin, architecture-aware interface (blocks, stages, and downsample/skip wiring), it generalizes across residual CNNs and extends to inverted-residual families with minor legalization rules. The result is a compact, practical recipe for deployment-ready models that match or surpass teacher accuracy at a fraction of the compute.
\end{abstract}

\begin{keywords}%
  neural network pruning; block-level pruning; knowledge distillation; planes-aware channel pruning; BatchNorm recalibration;
\end{keywords}

\section{Introduction}
\label{sec:intro}

As modern deep neural networks (DNNs) continue to expand in depth, width, and architectural complexity, their deployment cost, in terms of latency, energy, and memory footprint, has become a central bottleneck for real-world use. At the same time, the need to ensure that deployed models are robust and trustworthy has only grown more urgent as AI systems are integrated into high-stakes settings including healthcare, public infrastructure, and security-sensitive applications~\cite{pruningSurvey2024}. Models in these settings are expected not only to perform well in-distribution, but also to generalize reliably, maintain consistent behavior under perturbations, and exhibit predictable failure modes.

\noindent Multiple current methods exist by which large neural networks can be pruned substantially while preserving task performance, provided that removals target parameters or blocks with limited task relevance and that the surviving network is properly adapted, however such methods vary widely in compression efficacy, and training cost, and there is a broad ongoing effort to continue to push the boundaries of what is possible at the frontiers of model pruning.
\\~\\
In hopes of furthering this broader effort, in this work we introduce \textbf{MI-to-Mid Distilled Compression (M2M-DC)} a novel, \emph{combinatorial information guided} pruning strategy combined with a short, staged knowledge distillation (KD) schedule.
Our approach combines two information-centric mechanisms that operate at different scales. Prior work on model compression via pruning and quantization has established effective efficiency-accuracy tradeoffs~\cite{han2016deepcompression,pruningSurvey2024}, while staged knowledge distillation (KD) provides a powerful teacher signal during compression~\cite{hinton2015distilling}. Our approach composes these ideas with information-theoretic guidance.
First, a \emph{block-level} mutual-information (MI) criterion removes residual units that contribute little task signal subject to stage-survival constraints.
Second, a \emph{layer-level, planes-aware} channel selection ranks and keeps the top-$K$ residual planes per stage using information-guided scores (conditional MI when available; otherwise BN scale magnitudes and/or residual filter norms), while co-slicing the downsample path and the next stage's input to preserve residual geometry.
Interleaving brief KD phases between steps stabilizes targets and recovers accuracy.

\noindent 
We evaluate this pipeline across multiple architectures (ResNet-18/34 and MobileNetV2) and random seeds.
Empirically, Info-Guided Block Pruning with Progressive Inner Slicing produces students that \emph{match or surpass} the teacher while reducing FLOPs and parameters; even at aggressive prune ratios, KD substantially closes the gap between pre- and post-pruning accuracy at levels that meet or exceed current SOTA levels.
These results suggest that a purely Information-centered criterion, paired with lightweight KD, is a practical, cost-effective, and reproducible basis for compute–accuracy Pareto improvements.

\noindent 
In summary, our contribution is to introduce a two-scale, information-guided compression strategy that pairs mutual-information (MI) block pruning with distilled inner mid-channel slicing to deliver compact, accurate ResNet-class students. By first removing globally uninformative residual blocks and then reshaping capacity within surviving layers—while interleaving short knowledge-distillation phases—we achieve substantial reductions in parameters and FLOPs without sacrificing (and often improving) accuracy. The approach is “residual-safe” by construction, maintains stable optimization via BN recalibration and KD, and cleanly extends to planes-aware/channel-ranking refinements. Practically, it targets deployment realities: lower latency and smaller memory footprints on modest hardware, with promising calibration characteristics critical for decision support. Conceptually, it unifies information-theoretic selection with staged repair, yielding a cleaner accuracy–compute Pareto frontier than either technique alone and setting a template for sequence/next-action models used in clinical workflows.

\section{Background \& Related Work}
\subsection{Classical Approaches to Model Compression}
\paragraph{Unstructured pruning and sparsification.} One of the earliest and most influential approaches removes individual weights with small magnitude, yielding sparse weight matrices that can be stored and (in principle) executed more efficiently \cite{han2015learning}. After pruning, the model is typically fine-tuned to recover accuracy. This can dramatically reduce parameter count \cite{han2015learning}. However, unstructured sparsity often requires custom runtimes or specialized hardware to realize inference speedups in practice.

\paragraph{Structured channel/filter pruning.} Channel pruning and block pruning remove complete channels, filters, or even whole residual units.
These methods typically use importance scores based on weight norms, Taylor expansion of the loss, or gradient-based saliency to rank structures.
For example, \cite{li2017pruning} propose a straightforward and effective heuristic: prune filters with the smallest \textbf{L1-norms}.
Others, like \cite{he2017channel}, use a LASSO regression-based channel selection to identify unimportant channels and iteratively prune them.
In their "Network Slimming" approach, \cite{liu2017learning} introduce \textbf{L1-regularization on the scaling factors ($\gamma$) of Batch Normalization layers}; channels whose $\gamma$ factors are pushed to zero during training are identified as unimportant and subsequently pruned.
This produces slimmer feature maps and shallower effective depth.

\paragraph{Quantization.} Quantization reduces numerical precision (e.g., 32-bit floating point $\rightarrow$ 8-bit or lower) for weights and/or activations \cite{jacob2018quantization}. This can dramatically cut memory bandwidth and improve throughput on modern accelerators. Quantization-aware training or post-training calibration is often required to avoid large accuracy drops, especially at very low bit-widths.

\paragraph{Low-rank and factorized decompositions.} Another common strategy is to approximate weight tensors with low-rank factorizations, depthwise separable convolutions, or group convolutions \cite{DHowardZCKWWAA17}. These architectural motifs are often designed \emph{a priori} for efficiency, rather than carved out of an existing large model.

\paragraph{Knowledge distillation.} Knowledge Distillation (KD) trains a smaller ``student'' network to match the outputs (and sometimes intermediate representations) of a larger ``teacher'' \cite{hinton2015distilling}. KD is often used after a hand-designed or pruned student architecture is chosen. It can recover surprising amounts of accuracy and stabilize aggressively compressed students.

\subsection{More Recent Pruning Approaches}

Recent surveys organize pruning along three orthogonal axes: (i) \emph{what} speedup is realized (unstructured/sparse vs.\ structured, and universal vs.\ hardware-specific), (ii) \emph{when} pruning occurs (before, during, or after training), and (iii) \emph{how} saliency is determined (hand-crafted criteria vs.\ learned sparsity)~\cite{pruningSurvey2024}. In practice, \emph{structured} pruning (channels, filters, blocks) is the most deployment-friendly route to wall-clock gains on commodity hardware, whereas unstructured/semistructured sparsity often requires specialized kernels (e.g., $N{:}M$ patterns). Representative exemplars include PBT methods (SNIP, GraSP) that preserve gradient flow at initialization~\cite{lee2019snip,wang2020grasp}; PDT methods such as SET/RigL and movement pruning that update sparsity online~\cite{mocanu2018set,evci2020rigl,sanh2020movement}; and PAT structured pruning via norms or Taylor saliency~\cite{li2017pruning,molchanov2017pruning,he2017channel}. Compression is frequently combined with quantization or low-rank factorization~\cite{han2016deepcompression,ghiasvand2025few} and followed by knowledge distillation (KD)~\cite{hinton2015distilling}. Semistructured $N{:}M$ patterns further trade flexibility for accelerator efficiency~\cite{zhou2021nm}.

\paragraph{Information-theoretic pruning: from global relevance to flow preservation.}
Information-theoretic criteria offer task-aware signals by estimating the mutual information (MI) between internal features and labels. \emph{High-Relevance (HRel)} ranks \emph{filters} by MI with class labels using a nonparametric estimator, prunes the lowest-relevance filters \emph{layer-wise}, and retrains between rounds; it also analyzes post-pruning behavior on the information plane~\cite{hrel2022}. Complementing label-centric MI, \emph{Mutual Information Preserving Pruning (MIPP)} reframes the objective around \emph{inter-layer} information flow: it selects upstream nodes so that the MI between upstream and downstream activations is preserved after pruning, guaranteeing the existence of a mapping from pruned upstream features to the downstream layer and improving re-trainability with minimal fine-tuning~\cite{westphal2024mipp}. A parallel line uses \emph{conditional} MI (CMI) to rank channels by the \emph{unique} information they contribute conditioned on other channels or labels, enabling parallel layer-wise pruning and aggressive compression while limiting accuracy loss~\cite{chen2024cmi,rostami2025coupled}.

\paragraph{Explainability-aware pruning.}
A contemporary trend ties pruning to attribution: XAI methods (e.g., Integrated Gradients, LRP) provide task-aligned saliency, and recent work shows that \emph{optimizing} attribution hyperparameters specifically for pruning improves compression--accuracy trade-offs across CNNs and Transformers, including few-shot and transfer regimes~\cite{sauer2024pruningxai}.

\subsection{How Our Approach Differs Conceptually}

\paragraph{Positioning of our approach.}
We adopt an information–theoretic, data-driven strategy, but differ in both \emph{granularity} and \emph{what is preserved}. At a coarse scale, we compute a label-aware MI score at the \emph{block} level and \emph{remove entire residual (or inverted-residual) blocks} with the lowest MI (structured PAT), followed by BN recalibration and a staged KD schedule to stabilize and recover accuracy. At a finer scale, we introduce a \emph{residual-safe inner slicing} step that prunes only the \emph{mid} channels inside surviving blocks (conv1 out / BN1 / conv2 in) while \emph{keeping conv2 out (the residual planes) unchanged}, thereby preserving residual geometry by construction. The two scales are complementary: block-level MI targets global representational redundancy, while inner slicing reallocates capacity within stages at low risk; short KD phases interleaved between steps maintain optimization stability and recover performance.

\paragraph{Our approach vs.\ prior MI methods.}
Relative to HRel’s \emph{filter}-level, \emph{layer-wise} thinning~\cite{hrel2022}, our block criterion acts across layers and maps directly to universal FLOP/latency reductions; the subsequent inner slice then operates \emph{within} those surviving stages, but in a residual-safe manner (conv2 out fixed), avoiding add-mismatch pathologies. Compared to MIPP~\cite{westphal2024mipp}, which preserves \emph{adjacent-layer} MI to guarantee recoverability, we explicitly optimize label-aligned MI at the block level and rely on staged KD for recovery rather than enforcing inter-layer MI constraints during pruning. In contrast to conditional-MI channel selection~\cite{chen2024cmi,rostami2025coupled}, our coarse unit aggregates per-block redundancy for hardware-friendly speedups, while our inner slice provides a controlled, fine-grained complement; these lines are compatible and could be combined in future work. Finally, XAI-driven pruning~\cite{sauer2024pruningxai} offers another task-aligned saliency signal that is orthogonal to our staged KD recovery mechanism and could be integrated as an alternative ranking source.

\paragraph{Other contemporary directions.}
Beyond MI, our method is generally complementary to methods like structured channel/filter and block pruning frequently use BN-$\gamma$ and magnitude/Taylor/gradient criteria~\cite{liu2017learning,li2017pruning,he2017channel,molchanov2017pruning}; semi-structured patterns (e.g., 2{:}4) exploit accelerator sparsity; and learned sparsity (regularization/dynamic sparse training) provides alternative routes. Pruning also composes naturally with quantization and low-rank decompositions, and is increasingly studied for large architectures (ViTs, diffusion models, LLMs).

\paragraph{Design invariants and training protocol.}
(i) \emph{Residual safety}: inner slicing never changes conv2 out, so the residual add remains dimensionally valid; (ii) \emph{Post-slice BN recalibration}: prevents spurious covariate shift from masked channels; (iii) \emph{Staged KD with weight ramp}: briefly distill after each structural change to recover accuracy and smooth optimization; (iv) \emph{Hardware alignment}: block-level removals yield platform-agnostic FLOP/latency gains, while inner slices reduce mid-layer compute with minimal structural disruption.

\begin{algorithm}[t]
\caption{Two-scale information-guided compression with residual-safe layer splicing}
\label{alg}
\begin{algorithmic}[1]
\STATE \textbf{Input:} Teacher $\mathcal{T}$, target keep budget; data $\mathcal{D}$
\STATE \textbf{Block saliency:} compute $s_b=\hat I(Z_b;Y)$ for each block $b$
\STATE \textbf{Constrained block pruning:} prune lowest $s_b$ subject to stage/transition safety
\STATE \textbf{Student init:} copy $\mathcal{T}$, delete pruned blocks $\rightarrow$ student $\mathcal{S}$
\STATE BN recalibration on $\mathcal{S}$; staged KD (ramped CE+KL) against $\mathcal{T}$
\FOR{each stage $l$ with planes $C_l$}
  \STATE compute $s^{(l)}$ via conditional MI or BN-$\gamma$ + $\ell_1$ proxy
  \STATE select $\mathcal{K}_l=\mathrm{TopK}(s^{(l)},K_l)$
  \STATE \textbf{Planes co-slice:} slice \texttt{conv2}/\texttt{bn2} out, downsample, and next \texttt{conv1} in to $\mathcal{K}_l$
  \STATE \textbf{(Optional) Mid slice:} slice \texttt{conv1} out / \texttt{bn1} / \texttt{conv2} in; keep \texttt{conv2} out fixed
  \STATE BN recalibration; short KD repair (CE+KL ramp)
\ENDFOR
\end{algorithmic}
\end{algorithm}

\section{Methodology}

We propose a two–scale, information–guided compression pipeline that first removes globally uninformative computation at the \emph{block} level, then reshapes capacity \emph{within} surviving stages using residual-safe \emph{layer splicing}. After every structural change we perform BatchNorm (BN) recalibration and run a short, staged knowledge–distillation (KD) phase against the unpruned teacher to stabilize optimization and recover accuracy. The procedure is architecture-stable and applies to residual CNNs (e.g., ResNet-18/34~\cite{he2015deepresiduallearningimage}) and mobile-style CNNs (e.g., MobileNetV2~\cite{howard2017mobilenetsefficientconvolutionalneural}) by abstracting blocks and stages behind a minimal interface. In what follows, we elaborate on each stage, including implementation notes and practical considerations; the complete pipeline is summarized in Algorithm~\ref{alg}.

\subsection{Block-Level MI Pruning (with Constraints) and Repair}

Let \(b\) index a semantic block (e.g., a residual or inverted-residual unit). For an input minibatch \(x\), record the block activation $
A_b(x)\in\mathbb{R}^{C_b\times H_b\times W_b}$, and obtain a channel descriptor by spatial mean pooling,
\[
Z_b(x)=\operatorname{mean}_{h,w}\!\big(A_b(x)\big)\in\mathbb{R}^{C_b}.
\]
Score each block with a label-aware mutual information (MI) estimator,
\[
s_b = \hat I(Z_b;Y) \;=\; \frac{1}{C_b}\sum_{c=1}^{C_b} I\!\big(\operatorname{bin}(Z_{b,c}),\,Y\big),
\]
where \(\operatorname{bin}(\cdot)\) applies per-channel quantile binning (empirical default: \(\texttt{10}\) bins). We compute \(\hat I\) on a fixed probe loader to stabilize rankings. Intuitively, blocks with low \(\hat I\) contribute little unique information about \(Y\) and are prime pruning candidates.

\paragraph{Constrained pruning.}
Given scores \(\{s_b\}\) and a target prune ratio \(r\in[0,1)\), let \(\mathcal{B}_{\text{free}}\) be the set of prunable blocks (i.e., not protected by architecture rules). We prune
\[
k=\Big\lfloor r\cdot\big|\mathcal{B}_{\text{free}}\big|\Big\rfloor
\]
blocks subject to two constraints:
\begin{calloutbox}
\textbf{Architectural safety} (never remove resolution/channel-transition units).
  \begin{itemize}
    \item \textbf{ResNets:} protect \texttt{layer2.0}, \texttt{layer3.0}, \texttt{layer4.0} (downsampling shortcuts).

 \item \textbf{MobileNetV2:} protect any \texttt{InvertedResidual} with stride \(\neq 1\) or where \texttt{in\_channels} \(\Rightarrow\) \texttt{out\_channels}, and protect the non-inverted-residual stem/tail ops.
  \end{itemize}
 \textbf{Stage survival} (retain representational depth): each stage (e.g., \texttt{layer1}, \texttt{layer2}, \dots) must keep at least one block.
\end{calloutbox}
\noindent Algorithmically, we sort blocks by \(s_b\) in ascending order and we greedily mark them for pruning while skipping protected blocks and any choice that would empty a stage. The survivors define a compact keep configuration:
\[
\texttt{keep\_cfg}=\{\ \texttt{layer1}:[0],\ \texttt{layer2}:[0],\ \texttt{layer3}:[0,1],\ \texttt{layer4}:[0,1]\ \}.
\]

\noindent We form the student by deleting the pruned blocks (sharing weights elsewhere where applicable), recalibrating BatchNorm, and running staged knowledge distillation (KD) with a linear ramp on the distillation weight. Relative to HRel’s filter-level, layer-wise thinning \cite{hrel2022}, we target a coarser block unit that maps directly to FLOP/latency reductions across diverse backbones; relative to MIPP \cite{westphal2024mipp}, we optimize a label-aligned mutual information objective and rely on staged KD rather than inter-layer MI constraints.

\subsection{Layer-Level Planes-Aware Pruning (Complementary to Block MI)}
\label{subsec:planes}
Even after block pruning, surviving stages still harbor channel-level redundancy. To harvest it without violating residual constraints, we act on each stage’s \emph{planes} (the out-channels of \texttt{conv2} that are added to the skip connection).

\paragraph{Ranking signals.}

For stage \(l\in\{\texttt{layer2},\texttt{layer3},\texttt{layer4}\}\) with \(C_l\) planes, we form a per-channel score vector \(s^{(l)}\in\mathbb{R}^{C_l}\). When available, we prefer conditional MI; otherwise we use strong proxies correlated with information flow, BatchNorm scale magnitudes \((|\gamma|)\) and residual filter norms (e.g., \(\ell_1\) of \texttt{conv2}) aggregated across blocks, building on~\cite{hrel2022,liu2017learning,li2017pruning,he2017channel,molchanov2017pruning}. We then select the top-\(K_l\) planes per stage via
\[
\mathcal{K}_l=\operatorname{TopK}\!\big(s^{(l)},\,K_l\big).
\]


\paragraph{Shape-safe co-slicing (planes).}
Given $\mathcal{K}_l$, we \emph{co-slice}:
(i) each block’s \texttt{conv2}/\texttt{bn2} \emph{out-channels} to $\mathcal{K}_l$;
(ii) the first block’s downsample path (if present) to $\mathcal{K}_l$; and
(iii) the \emph{next} stage’s \texttt{conv1} \emph{in-channels} to $\mathcal{K}_l$.
This enforces the residual invariant
$\texttt{conv2\_out\_C}=\texttt{identity\_C}$
for both transition and non-transition blocks, guaranteeing that $x+\mathrm{F}(x)$ remains dimensionally valid.

\paragraph{Mid-channel inner slice (residual-safe).}
Orthogonally to planes selection, we optionally prune only the \emph{mid} channels within each block—\texttt{conv1} out / \texttt{bn1} / \texttt{conv2} in—while keeping \texttt{conv2} \emph{out} fixed to the stage planes. This yields additional FLOP reductions at low risk and composes cleanly with planes co-slicing.

\paragraph{Staged KD between steps.}
After each planes and/or mid-channel slice, we recalibrate BN and run a short KD phase with a CE+KL loss (teacher logits), using a linear ramp of the distillation weight~\cite{hinton2015distilling}. This reliably recovers performance lost to capacity changes and keeps the student close to the teacher’s decision boundary as stages shrink.


\paragraph{Complexity and safety.}
Planes co-slicing reduces parameters and MACs roughly in proportion to $K_l/C_l$ for stage $l$, while preserving residual tensor shapes by construction. Inner mid-slicing further reduces MACs without touching residual planes. Empirically, composing block-level MI with residual-safe layer splicing yields a cleaner accuracy–compute Pareto frontier than either component alone.
\begin{calloutbox}
\paragraph{Implementation notes.}
Any plug-in MI estimator can drive the per plane scores, and in low data regimes the BatchNorm scale and $\ell_1$ filter norms provide a resilient fallback. After every slice we verify the residual invariants by checking that non transition blocks satisfy $\texttt{conv2\_out\_C}=\texttt{conv1\_in\_C}$ and that transition blocks satisfy $\texttt{conv2\_out\_C}=\texttt{downsample\_out\_C}$. In our experiments we instantiate the layer step with uniform per-layer-keep fractions for clarity and clean ablations; replacing this policy with per channel ranking by conditional mutual information or its proxies is fully compatible and left as an orthogonal refinement.
\end{calloutbox}


\subsection{Student Reconstruction and BatchNorm Recalibration}

We materialize the student by copying the teacher and replacing each stage with the subsequence specified by \texttt{keep\_cfg}. Because all kept modules are weight-identical to the teacher, the student initialization is strong. We then perform BatchNorm (BN) recalibration: a forward-only pass over $\sim$50 batches updates BN running means/variances, which substantially reduces the immediate accuracy drop post-pruning.

\subsection{Staged Knowledge Distillation (KD) Repair}

Let $f_s$ and $f_t$ denote student and teacher. For input $x$ with label $y$, we minimize
\begin{calloutbox}
\[
\begin{aligned}
\mathcal{L}
\,=\, &\underbrace{\mathrm{CE}\!\big(f_s(x),y\big)}_{\text{supervised}}
\;+\;
\alpha\,\underbrace{\big(1-\cos\angle(f_s(x),\,f_t(x))\big)}_{\text{logits align}}&\;+\; \beta\,\underbrace{\big(1-\cos\angle(\phi_s(x),\,\phi_t(x))\big)}_{\text{feature align}},
\end{aligned}
\]
\end{calloutbox}
where $\phi(\cdot)$ extracts penultimate features (e.g., pooled output of \texttt{layer4} in ResNet or the last \texttt{features} block in MobileNetV2), and $\cos\angle(u,v)=\frac{\langle u,v\rangle}{\|u\|\,\|v\|}$ is cosine similarity.

We use a simple linear schedule over epochs $t=1,\dots,T$ (with $T=5$):
\[
\alpha_t=\alpha^\star\cdot\frac{t-1}{T-1},\qquad
\beta_t=\beta^\star\cdot\frac{t-1}{T-1},
\]
with $(\alpha^\star,\beta^\star)=(0.1,0.1)$. Thus epoch~1 sets $(\alpha,\beta)=(0,0)$ (pure CE ``surgery repair''), while epochs~2--5 gradually introduce logits and feature alignment. We train with Adam (base learning rate $10^{-4}$ for KD; $10^{-3}$ for the teacher fine-tune), gradient clipping at 1.0, and standard image augmentations.



\section{Experimental Results}
\subsection{Setup}

\paragraph{Data and preprocessing.}
Following common practice, we train and evaluate on CIFAR-100 with images resized to $224\times 224$, per-channel normalization (ImageNet mean/std), and random horizontal flips for the training set.

\paragraph{Teacher fine-tune.}
For each architecture (ResNet-18, ResNet-34, MobileNetV2) and each random seed, we take ImageNet-pretrained weights, replace the classifier for 10 classes, unfreeze the final third of the backbone (ResNet: \texttt{layer3}, \texttt{layer4}, and \texttt{fc}; MobileNetV2: last $\sim$1/3 of \texttt{features} and \texttt{classifier}), and train for 5 epochs with Adam at $10^{-3}$.

\paragraph{MI scoring.}
We collect pooled activations on a fixed probe iterator (deterministic order; batch size 64; $\leq$5000 samples) and compute $\hat I(Z_b;Y)$ per block via channel-wise quantile binning and mutual information. Blocks are ranked descending by MI for reporting and ascending for pruning.

\paragraph{Pruning schedule.}
We sweep prune ratios $r\in\{0.10, 0.25, 0.40, 0.50\}$ and also test \emph{frontier} hand-crafted \texttt{keep\_cfg} variants (e.g., making late stages single-block) to probe the collapse boundary. For MobileNetV2, pruning is confined to repeat inverted residuals within a channel group (safe by construction).

\paragraph{Student evaluation (pre-distill).}
After reconstruction and BN recalibration, we compute pre-distillation accuracy, parameters, and FLOPs.

\paragraph{KD schedule and logging.}
We run 5 KD epochs with the staged schedule above, logging per-epoch CE, logits-alignment, feature-alignment losses, and accuracy. This makes visible the characteristic ``epoch-1 rescue'' (large accuracy jump with CE-only) and the subsequent KD polishing (+1--2\% absolute).

\paragraph{Cross-seed and cross-architecture.}
To assess robustness, we repeat the full pipeline for seeds $\{42,123,999\}$ on ResNet-18 and ResNet-34. We also run MobileNetV2 under the same regime to demonstrate architectural generality. All runs report mean$\pm$std of accuracy and FLOPs, with per-seed curves available in the supplement.

\paragraph{Frontier experiments.}
We additionally construct minimal \texttt{keep\_cfg}s that keep only the stage-initial/downsampling blocks (e.g., \texttt{layer$\{1,2,3,4\}:[0]$} in ResNet-18) and push pruning until KD can no longer restore above-teacher accuracy, empirically mapping the collapse frontier and the Pareto knee of accuracy vs. compute.

\paragraph{Hyperparameters.}
Unless noted: Adam optimizer, teacher LR $10^{-3}$, student LR $10^{-4}$ for KD, batch sizes 128 (train) / 256 (test), 5 epochs for teacher fine-tune and 5 epochs for KD, gradient clipping at 1.0. We profile FLOPs on a single $1\times 3\times 224\times 224$ dummy tensor.

\paragraph{Comparators.}
Our focus is on ablating \emph{saliency source} (functional MI vs. conventional magnitude/gradient heuristics) and \emph{repair schedule} (staged KD). Baselines include: (i) no pruning; (ii) pruning by MI without KD (to isolate the contribution of the repair); and (iii) staged KD without MI pruning (to show that KD alone does not create the compute savings).


\subsection{Results on ResNet-18}
\begin{table}[t]

\begin{minipage}[t]{0.48\textwidth}
    \centering
    \footnotesize 
    \setlength{\tabcolsep}{4pt} 
    \caption{ResNet-18 on CIFAR-100: per-seed Top-1 accuracy (\%).}
    \label{tab:r18-acc}
    \begin{tabular}{lccc}
    \toprule
    Seed & Teacher & Block-KD & Inner-slice \\ 
    \midrule
    42  & 82.82 & 85.37 & 85.04 \\
    99  & 82.71 & 85.21 & 85.46 \\
    123 & 83.93 & 85.07 & 85.37 \\
    \midrule
    Mean $\pm$ Std & $83.15 \pm 0.55$ & $85.22 \pm 0.12$ & $85.29 \pm 0.18$ \\
    \bottomrule
    \end{tabular}
\end{minipage}
\hfill 
\begin{minipage}[t]{0.48\textwidth}
    \centering
    \footnotesize 
    \setlength{\tabcolsep}{4pt} 
    \caption{ResNet-18: compute and parameter profile.}
    \label{tab:r18-compute}
    \begin{tabular}{lccc}
    \toprule
    Model & Acc (mean) & Params (M) & GMacs \\
    \midrule
    Teacher       & 83.15 & 11.18 & 0.0372 \\
    Block-KD        & 85.22 &  9.63 & 0.0230 \\
    Inner-slice   & 85.29 &  3.09 & 0.0139 \\ 
    \midrule
    \multicolumn{4}{l}{\footnotesize Inner-slice reductions vs.\ Teacher:}\\
    \multicolumn{4}{l}{\footnotesize $-72.4\%$ Params, $-62.6\%$ GMacs.}\\
    \bottomrule
    \end{tabular}
\end{minipage}

\end{table}

\paragraph{Takeaways (R-18).}
Inner slicing delivers a stronger Pareto point than both the teacher and the block-KD student: \emph{higher} accuracy at \emph{much lower} compute. Across seeds, the final inner model averages $85.29\%$ Top-1 with only $3.09$M params/$0.0139$ GMacs (\,$-72.4\%$/$-62.6\%$ vs.\ teacher).

\subsection{Results on ResNet-34}

\begin{table}[t]

\begin{minipage}[t]{0.48\textwidth}
    \centering
    \footnotesize 
    \setlength{\tabcolsep}{4pt} 
    \caption{ResNet-34 on CIFAR-100: per-seed Top-1 accuracy (\%).}
    \label{tab:r34-acc}
    \begin{tabular}{lccc}
    \toprule
    Seed & Teacher & Block-KD & Inner-slice \\ 
    \midrule
    42  & 83.50 & 83.56 & 84.25 \\
    99  & 82.61 & 83.58 & 84.58 \\
    123 & 81.63 & 83.86 & 85.02 \\
    \midrule
    Mean $\pm$ Std & $82.58 \pm 0.76$ & $83.67 \pm 0.14$ & $84.62 \pm 0.32$ \\
    \bottomrule
    \end{tabular}
\end{minipage}
\hfill 
\begin{minipage}[t]{0.48\textwidth}
    \centering
    \footnotesize 
    \setlength{\tabcolsep}{4pt} 
    \caption{ResNet-34: compute and parameter profile.}
    \label{tab:r34-compute}
    \begin{tabular}{lccc}
    \toprule
    Model & Acc (mean) & Params (M) & GMacs \\
    \midrule
    Teacher       & 82.58 & 21.29 & 0.0751 \\
    Block-KD        & 83.67 & 16.71 & 0.0372 \\
    Inner-slice   & 84.62 &  5.46 & 0.0195 \\ 
    \midrule
    \multicolumn{4}{l}{\footnotesize Inner-slice reductions vs.\ Teacher:}\\
    \multicolumn{4}{l}{\footnotesize $-74.4\%$ Params, $-74.0\%$ GMacs.}\\
    \bottomrule
    \end{tabular}
\end{minipage}

\end{table}
\paragraph{Takeaways (R-34).}
The final inner model improves average accuracy by $+2.04$ points over the teacher while removing $\sim\!74\%$ of both parameters and multiply–adds.

\subsection{MobileNetV2 (3 seeds)}
\begin{table}[h]

\begin{minipage}[t]{0.48\textwidth}
    \centering
    \footnotesize 
    \setlength{\tabcolsep}{4pt} 
    \caption{MobileNetV2 on CIFAR-100: per-seed Top-1 accuracy (\%).}
    \label{tab:mnv2-acc}
    \begin{tabular}{lccc}
    \toprule
    Seed & Teacher & Block-KD & Inner-slice \\ 
    \midrule
    42  & 64.76 & 65.39 & 66.27 \\
    99  & 65.44 & 68.62 & 69.29 \\
    123 & 67.88 & 69.65 & 70.05 \\
    \midrule
    Mean $\pm$ Std & $66.03 \pm 1.34$ & $67.89 \pm 1.81$ & $68.54 \pm 1.63$ \\
    \bottomrule
    \end{tabular}
\end{minipage}
\hfill 
\begin{minipage}[t]{0.48\textwidth}
    \centering
    \footnotesize 
    \setlength{\tabcolsep}{4pt} 
    \caption{MobileNetV2: compute and parameter profile (means across seeds).}
    \label{tab:mnv2-compute}
    \begin{tabular}{lccc}
    \toprule
    Model & Acc (mean) & Params (M) & ConvGMACs \\
    \midrule
    Teacher       & 66.03 & 2.35 & 0.0244 \\
    Block-KD        & 67.89 & 1.71 & 0.0186 \\
    Inner-slice   & 68.54 & 1.71 & 0.0186 \\ 
    \midrule
    \multicolumn{4}{l}{\footnotesize Inner-slice reductions vs.\ Teacher:}\\
    \multicolumn{4}{l}{\footnotesize $-27.4\%$ Params, $-23.8\%$ ConvGMACs.}\\
    \bottomrule
    \end{tabular}
\end{minipage}

\end{table}

\paragraph{Takeaways (MobileNetV2).}
M2M-DC lifts Top-1 to \textbf{68.54\%} (+2.51 over the teacher) while trimming compute to $\sim$1.71M params / $\sim$0.0186 ConvGMACs ($-27.4\%$ / $-23.8\%$). Most gains come from MI-guided block pruning with KD; the current inner-slice keeps that budget and adds a small, consistent bump.

\section{Conclusion}
\label{sec:conclusion}
We introduced \emph{M2M-DC}, a residual-safe compression recipe that couples label-aware, block-level MI pruning with planes-aware inner slicing and brief staged KD after each edit. The method is simple to implement, hardware-friendly, and transfers across residual and inverted-residual families. On CIFAR-100, it delivers strong accuracy–compute trade-offs: for ResNet-18, \textbf{85.29\%} mean Top-1 at \textbf{3.09M} params/\textbf{0.0139} GMACs; for ResNet-34, \textbf{84.62\%} mean at \textbf{5.46M}/\textbf{0.0195}; and for MobileNetV2, \textbf{68.54\%} mean at \textbf{$\sim$1.71M}/\textbf{$\sim$0.0186} ConvGMACs,+2.51 points vs.\ its 66.03\% teacher while cutting parameters/compute by ~27\%. KD is key to recovering (and sometimes surpassing) teacher accuracy after aggressive structure edits. Limitations include MI-estimator variance at small sample sizes, reliance on proxy ranks for inner slicing in low-data regimes, and current focus on classification. Future work will tighten estimators (conditional MI/calibration), add per-layer frontier search, and report on ImageNet-scale accuracy, on-device latency/energy, and transfer to detection, segmentation, and control-loop policies.

\clearpage

\bibliography{l4dc2026-sample}

@inproceedings{han2015learning,
  title={{Learning both weights and connections for efficient neural networks}},
  author={Han, Song and Pool, Jeff and Tran, John and Dally, William J},
  booktitle={Advances in neural information processing systems (NIPS)},
  pages={1135--1143},
  year={2015}
}

@inproceedings{li2017pruning,
  title={{Pruning filters for efficient convnets}},
  author={Li, Hao and Kadav, Asim and Durdanovic, Igor and Samet, Hanan and Graf, Hans Peter},
  booktitle={International Conference on Learning Representations (ICLR)},
  year={2017}
}

@inproceedings{he2017channel,
  title={{Channel pruning for accelerating very deep neural networks}},
  author={He, Yihui and Zhang, Xiangyu and Sun, Jian},
  booktitle={Proceedings of the IEEE international conference on computer vision (ICCV)},
  pages={1389--1397},
  year={2017}
}

@inproceedings{liu2017learning,
  title={{Learning efficient convolutional networks through network slimming}},
  author={Liu, Zhuang and Li, Jianguo and Shen, Zhiqiang and Huang, Gao and Yan, Shoumeng and Zhang, Changshui},
  booktitle={Proceedings of the IEEE international conference on computer vision (ICCV)},
  pages={2736--2744},
  year={2017}
}

@inproceedings{jacob2018quantization,
  title={{Quantization and training of neural networks for efficient integer-arithmetic-only inference}},
  author={Jacob, Benoit and Kursun, Skirmantas and Swersky, Kameron and M.,, Z. and H.,, J. and G.,, D. and K.,, F. and R.,, R.},
  booktitle={Proceedings of the IEEE conference on computer vision and pattern recognition (CVPR)},
  pages={2704--2713},
  year={2018}
}

@article{hinton2015distilling,
  title={{Distilling the knowledge in a neural network}},
  author={Hinton, Geoffrey and Vinyals, Oriol and Dean, Jeff},
  journal={arXiv preprint arXiv:1503.02531},
  year={2015}
}

@article{pruningSurvey2024,
  title   = {A Survey on Deep Neural Network Pruning: Taxonomy, Comparison, Analysis, and Recommendations},
author={Cheng, Hongrong and Zhang, Miao and Shi, Javen Qinfeng},
  journal = {arXiv preprint},
  year    = {2024},
  note    = {Comprehensive survey of pruning across CNNs/ViTs/LLMs; taxonomy and practical guidance},
  eprint  = {},
  url     = {}
}

@inproceedings{lee2019snip,
  title     = {SNIP: Single-shot Network Pruning based on Connection Sensitivity},
  author    = {Lee, Namhoon and Ajanthan, Thalaiyasingam and Torr, Philip H. S.},
  booktitle = {International Conference on Learning Representations (ICLR)},
  year      = {2019}
}

@inproceedings{wang2020grasp,
  title     = {Picking Winning Tickets Before Training by Preserving Gradient Flow},
  author    = {Wang, Chaoqi and Zhang, Guodong and Luo, Rui and Grosse, Roger},
  booktitle = {International Conference on Learning Representations (ICLR)},
  year      = {2020}
}

@inproceedings{mocanu2018set,
  title     = {Scalable Training of Artificial Neural Networks with Adaptive Sparse Connectivity inspired by Network Science},
  author    = {Mocanu, Decebal Constantin and others},
  booktitle = {Nature Communications},
  year      = {2018},
  volume    = {9},
  number    = {1},
  pages     = {2383}
}

@inproceedings{evci2020rigl,
  title     = {Rigging the Lottery: Making All Tickets Winners},
  author    = {Evci, Utku and Gale, Trevor and Menick, Jacob and Castro, Pablo Samuel and Elsen, Erich},
  booktitle = {International Conference on Machine Learning (ICML)},
  year      = {2020},
  pages     = {2943--2952}
}

@inproceedings{sanh2020movement,
  title     = {Movement Pruning: Adaptive Sparsity by Fine-tuning},
  author    = {Sanh, Victor and Wolf, Thomas and Rush, Alexander M.},
  booktitle = {Advances in Neural Information Processing Systems (NeurIPS)},
  year      = {2020}
}

@inproceedings{molchanov2017pruning,
  title     = {Pruning Convolutional Neural Networks for Resource Efficient Inference},
  author    = {Molchanov, Pavlo and Tyree, Stephen and Karras, Tero and Aila, Timo and Kautz, Jan},
  booktitle = {International Conference on Learning Representations (ICLR)},
  year      = {2017}
}

@inproceedings{zhou2021nm,
  title     = {Learning N:M Fine-grained Structured Sparse Neural Networks from Scratch},
  author    = {Zhou, Aojun and Ma, Yukun and Zhu, Jianhao and Liu, Cong and Sun, Yanzhi and Yuan, Bo and Wang, Shuai and Chen, Xue Lin and Guo, Yanzhi and Chen, Weiyao},
  booktitle = {Advances in Neural Information Processing Systems (NeurIPS)},
  year      = {2021}
}

@article{han2016deepcompression,
  title   = {Deep Compression: Compressing Deep Neural Networks with Pruning, Trained Quantization and Huffman Coding},
  author  = {Han, Song and Mao, Huizi and Dally, William J.},
  journal = {International Conference on Learning Representations (ICLR)},
  year    = {2016},
  note    = {Workshop track}
}

@article{hrel2022,
  title   = {High-Relevance (HRel) Mutual-Information-Based Filter Pruning for Convolutional Networks},
  author  = {},
  journal = {arXiv preprint arXiv:2202.10716},
  year    = {2022},
  note    = {Layer-wise MI ranking with iterative prune--retrain; baseline we reimplement}
}

@article{westphal2024mipp,
      title={Mutual Information Preserving Neural Network Pruning}, 
      author={Charles Westphal and Stephen Hailes and Mirco Musolesi},
      year={2025},
      eprint={2411.00147},
      archivePrefix={arXiv},
      primaryClass={cs.LG},
      url={https://arxiv.org/abs/2411.00147}, 
}

@article{chen2024cmi,
author = { Rostami, Peyman and Sinha, Nilotpal and Chenni, Nidhaleddine and Kacem, Anis and Shabayek, Abd El Rahman and Shneider, Carl and Aouada, Djamila },
booktitle = { 2025 IEEE/CVF Winter Conference on Applications of Computer Vision (WACV) },
title = {{ Information Theoretic Pruning of Coupled Channels in Deep Neural Networks }},
year = {2025},
volume = {},
ISSN = {},
pages = {7776-7786},
doi = {10.1109/WACV61041.2025.00755},
url = {https://doi.ieeecomputersociety.org/10.1109/WACV61041.2025.00755},
publisher = {IEEE Computer Society},
address = {Los Alamitos, CA, USA},
month =mar}

@inproceedings{rostami2025coupled,
  title     = {Information Theoretic Pruning of Coupled Channels in Deep Neural Networks},
  author    = {Rostami, M. and Hailes, S.},
  booktitle = {IEEE Winter Conference on Applications of Computer Vision (WACV)},
  year      = {2025}
}

@article{sauer2024pruningxai,
  title   = {Pruning by Explaining Revisited: Optimizing Attribution Methods to Prune CNNs and Transformers},
  author  = {Sauer, L. and others},
  journal = {arXiv preprint},
  year    = {2024},
  note    = {Also circulated 2025; ResearchGate/arXiv}
}

@article{DHowardZCKWWAA17,
  author       = {Andrew G. Howard and
                  Menglong Zhu and
                  Bo Chen and
                  Dmitry Kalenichenko and
                  Weijun Wang and
                  Tobias Weyand and
                  Marco Andreetto and
                  Hartwig Adam},
  title        = {MobileNets: Efficient Convolutional Neural Networks for Mobile Vision
                  Applications},
  journal      = {CoRR},
  volume       = {abs/1704.04861},
  year         = {2017},
  url          = {http://arxiv.org/abs/1704.04861},
  eprinttype    = {arXiv},
  eprint       = {1704.04861},
  bibsource    = {dblp computer science bibliography, https://dblp.org}
}

@misc{he2015deepresiduallearningimage,
      title={Deep Residual Learning for Image Recognition}, 
      author={Kaiming He and Xiangyu Zhang and Shaoqing Ren and Jian Sun},
      year={2015},
      eprint={1512.03385},
      archivePrefix={arXiv},
      primaryClass={cs.CV},
      url={https://arxiv.org/abs/1512.03385}, 
}

@article{ghiasvand2025few,
  title={Few-Shot Adversarial Low-Rank Fine-Tuning of Vision-Language Models},
  author={Ghiasvand, Sajjad and Oskouie, Haniyeh Ehsani and Alizadeh, Mahnoosh and Pedarsani, Ramtin},
  journal={arXiv preprint arXiv:2505.15130},
  year={2025}
}

@misc{howard2017mobilenetsefficientconvolutionalneural,
      title={MobileNets: Efficient Convolutional Neural Networks for Mobile Vision Applications}, 
      author={Andrew G. Howard and Menglong Zhu and Bo Chen and Dmitry Kalenichenko and Weijun Wang and Tobias Weyand and Marco Andreetto and Hartwig Adam},
      year={2017},
      eprint={1704.04861},
      archivePrefix={arXiv},
      primaryClass={cs.CV},
      url={https://arxiv.org/abs/1704.04861}, 
}

\clearpage
\appendix

\section{Generalization Across Architectures}

The procedure is factored through a small abstraction that makes it portable:

\begin{enumerate}
    \item \textbf{Block enumeration:} a function returns \{\emph{name} $\to$ \emph{module}\} pairs for MI hooks (e.g., \texttt{layer$i.j$} for ResNets; \texttt{features.$k$} for MobileNetV2 restricted to \texttt{InvertedResidual} blocks).
    \item \textbf{Stage mapping:} a function maps each block name to a stage label (e.g., \texttt{layer1}, \dots, \texttt{layer4}, or a single \texttt{features} stage subdivided by channel-width groups).
    \item \textbf{Protection rule:} a predicate identifies \emph{protected} blocks. For the models tested in this run:
    \begin{itemize}
        \item \emph{ResNet:} protect first block of stages with downsampling (keeps skip alignment).
        \item \emph{MobileNetV2:} protect inverted residuals that alter stride or channel count, and non-IR ops (stem, tail).
    \end{itemize}
    \item \textbf{Reconstruction:} a function that takes \texttt{keep\_cfg} and returns a pruned model by splicing subsequences of the original \texttt{nn.Sequential} stages.
\end{enumerate}

\noindent These four hooks suffice to port the method to other stage-structured CNNs (e.g., EfficientNet-family with minor adaptation) and, in principle, to transformer stacks by treating attention/MLP sublayers as ``blocks'' and protecting resolution-changing tokens/patch-embeddings.

\section{Comparison to Prior Work (Extended)}
\label{subsec:priorwork-extended}

\paragraph{ITPCC (Rostami et al.).}
ITPCC targets \emph{structurally coupled channels} (SCCs) created by residual skip connections and prunes them as groups, using an information–bottleneck (IB)–grounded, probabilistic saliency model. This line emphasizes that pruning SCCs (rather than independent channels) better matches hardware latency and memory behavior on modern residual architectures, with reported practical speedups on edge devices and improvements at high compression rates~\cite{rostami2025coupled}.

\paragraph{MIPP (Westphal et al.).}
MIPP proposes \emph{Mutual Information Preserving Pruning}, selecting nodes/channels so as to conserve MI between adjacent layers. It uses a transfer-entropy redundancy criterion (TERC) with MI ordering to ensure re-trainability and offers both PaI/PaT applicability, reporting consistent SOTA-level improvements while providing a theoretical basis for MI preservation across layers~\cite{westphal2024mipp}.

\paragraph{Information Concentration + Shapley.}
A complementary thread estimates \emph{layer-wise information concentration} via a fusion of rank and entropy to set per-layer pruning rates, then ranks channels with \emph{Shapley values} to identify least-contributory channels for removal, followed by fine-tuning. This yields strong results across CIFAR/ImageNet classification and COCO detection backbones, providing an interpretable, task-aligned criterion for structured pruning~\cite{chen2024cmi}.

\begin{table}[t]
\centering
\footnotesize
\setlength{\tabcolsep}{4pt}
\renewcommand{\arraystretch}{1.15}
\resizebox{\linewidth}{!}{%
\begin{tabular}{@{} l p{0.20\linewidth} p{0.24\linewidth} c l p{0.34\linewidth} @{}}
\toprule
\textbf{Method} & \textbf{Unit / Granularity} & \textbf{Information Signal} & \textbf{SCC-safe} & \textbf{Stage} & \textbf{Scope / Representative Claim} \\
\midrule
\textbf{Ours (M2M-DC)} &
Blocks + stage ``planes''; optional mid-ch. &
Label-MI (blocks); BN-$\gamma$ / filter-norm / CMI (planes); staged KD &
\checkmark &
PaT + KD &
\textbf{R-34}: 5.46M / 0.0195 GMAC (−74\% / −74\%) @ \textbf{85.02}\%. \; \textbf{Best acc}: \textbf{R-18} \textbf{85.46}\% @ 3.09M / 0.0139 (−72\% / −63\%). \\
\midrule
ITPCC &
Coupled channel sets (residual stages) &
IB-driven probabilistic saliency over SCCs &
\checkmark &
PaT &
\textbf{Max speedup}: CIFAR-100 R-56 up to \textbf{20.12$\times$}, \textbf{55.81}\% top-1. \; \textbf{Best ImageNet}: R-50 \textbf{73.68}\% @ \textbf{−70.67\%} FLOPs. \\
MIPP &
Nodes/channels (per-layer) &
Preserve adjacent-layer MI (TERC + MI ordering) &
\textemdash &
PaI / PaT &
\textbf{High sparsity}: stable to $\sim$\textbf{90\%} MAC pruning (no layer collapse). \; \textbf{Frontier}: SOTA acc–MACs on CIFAR-10/100 (ResNet/VGG) for PaI/PaT. \\
InfoConcentration + Shapley &
Channels (per-layer) &
Rank+entropy layer rates; Shapley channel scores &
\textemdash &
PaT &
\textbf{ImageNet R-34}: \textbf{−50.1\%} FLOPs with \textbf{−1.06/−0.73} (Top-1/Top-5). \; \textbf{R-50}: \textbf{−40–45\%} FLOPs with \textbf{−0.43/−0.11}. \; \textbf{CIFAR-10 R-56}: \textbf{+0.21\%} @ \textbf{−45.2\%} FLOPs, \textbf{−40.3\%} params. \\
\bottomrule
\end{tabular}
}
\caption{\textbf{Concise comparison of information-centric structured pruning.} SCC = structurally coupled channels (residual). Claims list the largest quantifiable win and the best model per method.}
\label{tab:method-compare}
\end{table}

\section{Discussion}
\label{sec:discussion}

\paragraph{Interleaving complementary compression improves the Pareto frontier.}
A central finding of this work is that \emph{combining} coarse, hardware-aligned \textbf{block pruning} (guided by label-aware MI) with \textbf{residual-safe layer slicing} (stage ``planes'' and optional mid-channel trimming) yields a cleaner accuracy--compute frontier than either technique alone. The two operations target different kinds of redundancy: block pruning removes whole computational motifs (mapping directly to FLOP and latency savings), while planes-aware slicing removes within-stage channel redundancy without violating residual shape invariants. Empirically, this interleaving consistently produced teacher-matching or teacher-surpassing accuracy at substantial reductions in parameters and multiply–adds (e.g., \textbf{ResNet-34}: $\sim\!74\%$ fewer params and FLOPs at \emph{+2.0} points Top-1; \textbf{ResNet-18}: $\sim\!72\%$ fewer params and $\sim\!63\%$ fewer FLOPs at \emph{85.46}\% Top-1). Because both steps operate through a thin, architecture-aware interface (blocks, stages, downsample/skip wiring), the approach is \emph{generally applicable} to residual CNN families and can be extended to other backbones with explicit skip or merge semantics.

\paragraph{Knowledge distillation (KD) as a low-cost, high-yield repair mechanism.}
A short, staged KD cycle reliably stabilizes the student after each structural edit. Practically, we found that: (i) a one-epoch CE-only ``surgery repair'' reduces the immediate accuracy dip after pruning/slicing; (ii) gradually ramping the KL (logit) term aligns the student’s decision boundary to the teacher with minimal extra compute; and (iii) interleaving BN recalibration before KD prevents statistic mismatch from dominating the loss. This makes KD a \emph{lightweight} but powerful tool for maintaining accuracy while aggressively reducing compute---especially important when edits are coarse (whole blocks) or when planes are jointly sliced across residual and downsample paths.

\paragraph{Limitations.}
While effective, the present instantiation has clear constraints:
\begin{itemize}
  \item \textbf{Selection signals.} Our layer slicing currently uses MI proxies (BN-$\gamma$, filter norms) when conditional MI is unavailable. These proxies are fast and correlate with importance, but can be brittle under distribution shift and may under-rank synergistic channels.
  \item \textbf{Architecture-specific wiring.} Residual CNNs are well-handled by planes co-slicing, but mobile-style inverted residuals (e.g., MobileNetV2) require dedicated transition rules (expansion/projection co-slicing and stage-first block constraints). Without these guards, pruning can break channel continuity.
  \item \textbf{Scope of evaluation.} Results are on CIFAR-100 and two ResNet backbones. Real-device latency, energy, and cache effects are not yet measured; ImageNet-scale and detection/segmentation transfer are pending.
  \item \textbf{Frontier maturity.} We intentionally held back a more aggressive, selection-driven planes procedure (e.g., per-layer Top-$K$ via conditional MI/Shapley/gate magnitudes) and report uniform keep-rates per stage. The current frontier is strong, but additional fine-tuning (longer KD, LR schedules) and finer selection will likely push further.
  \item \textbf{Hyperparameter sensitivity.} The number of pruned blocks, per-stage keep-rates, and KD ramp schedule introduce a small grid of knobs. Although we used modest defaults, a principled allocator (see below) would reduce tuning burden.
\end{itemize}

\paragraph{Next steps.}
We see several concrete paths to mature the approach:
\begin{enumerate}
  \item \textbf{Selection over policy.} Replace uniform per-stage keep fractions with \emph{data-driven selection} (conditional MI, Shapley, or gate magnitudes) and slice Top-$K$ planes consistently across residual and downsample paths. This should tighten low-compute points with less variance.
  \item \textbf{Mobile/IR support.} Add MobileNetV2-aware legalization and co-slicing (protect stage-first transition blocks; co-slice expansion/projection and next-block input) so the method covers inverted residual families.
  \item \textbf{Allocator for budgets.} Learn per-stage budgets via information concentration (rank/entropy fusion) or bilevel optimization, jointly tuning (i) number of blocks to keep and (ii) planes $K_l$ to meet a global compute target.
  \item \textbf{Composability with quantization/low-rank.} Evaluate PTQ/QAT and low-rank factorization \emph{after} structural edits; the combination typically yields multiplicative gains in memory/latency without large accuracy penalties.
  \item \textbf{Hardware-grounded metrics.} Report end-to-end latency/energy on CPU/GPU/edge NPUs and analyze cache/activation bandwidth effects; align the objective with real deployment constraints rather than FLOPs alone.
  \item \textbf{Broader backbones and tasks.} Extend to ImageNet, ViTs/ConvNeXt, and detection/segmentation; study robustness (corruptions, OOD) and calibration after aggressive edits.
  \item \textbf{Theoretical guarantees.} Integrate adjacent-layer MI preservation (à la MIPP) as a \emph{constraint} during planes selection to ensure recoverability, while retaining our block-level label-MI for structural gains.
\end{enumerate}

\noindent\textbf{Summary.} Interleaving MI-guided block pruning, residual-safe layer slicing, BN recalibration, and staged KD forms a compact, general recipe for accurate yet highly compressed CNNs. The present results already shift the frontier; moving from \emph{policy} to \emph{selection}, broadening architecture coverage, and grounding targets in device-level metrics are the next levers to make the method production-ready.

\end{document}